\pdfoutput=1

\documentclass[11pt]{article}

\usepackage{acl}
\usepackage{times}
\usepackage{latexsym}

\usepackage[T1]{fontenc}

\usepackage[utf8]{inputenc}

\usepackage{microtype}

\usepackage[linesnumbered,ruled,vlined]{algorithm2e}
\usepackage{array}
\usepackage{colortbl}
\usepackage{subcaption}
\usepackage{booktabs}
\usepackage{amsfonts}
\usepackage{float}
\usepackage{todonotes}
\usepackage{soul}
\usepackage{multirow, multicol}

\usepackage{newfloat}
\usepackage{listings}
\usepackage{pifont}
\usepackage{breqn}
\usepackage[capitalize,nameinlink]{cleveref}

\newcommand{\cmark}{\ding{51}}%
\newcommand{\MDF}{Most Dominant Frames}
\newcommand{\MIF}{Most Implied Frames}

\DeclareMathOperator*{\argtopk}{arg\,topk}

\title{Self-Adaptive Sampling for Efficient Video Question Answering on Image--Text Models}

\author{
    Wei Han \textsuperscript{$\dagger$}\thanks{~~Corresponding authors: wei\_han@mymail.sutd.edu.sg, hui\_chen@mymail.sutd.edu.sg} \quad
    Hui Chen \textsuperscript{$\dagger$} \quad
    Min-Yen Kan \textsuperscript{$\ddagger$} \quad
    Soujanya Poria \textsuperscript{$\dagger$} \\
    \textsuperscript{$\dagger$} Singapore University of Technology and Design, \textsuperscript{$\ddagger$} National University of Singapore \\
}

\begin{document}
\maketitle
\begin{abstract}
    Image--text models (ITMs) are the prevalent architecture to solve video question--answering tasks. ITMs requires only a few input frames, saving significant computation over against video--language models.
    However, we find existing ITM video question--answering either 1) adopts simplistic and unintentional sampling strategies, which may miss key frames that offer answer clues; or 2) samples a large number of frames into divided groups, which computational sources can not accommodate. 
    We develop an efficient sampling method for the few-frame scenario.
    We first summarize a family of prior sampling methods based on question--frame correlation into a unified one, dubbed~\textit{\MIF}~(MIF). Through analysis, we form a  hypothesis that question-aware sampling is not necessary, from which we further propose the second method~\textit{\MDF}~(MDF).
    Results on four public datasets and three ITMs demonstrate that MIF and MDF boost the performance for image--text pretrained models, and have a wide application over both model architectures and datasets. Code is available at \url{https://github.com/declare-lab/Sealing}.
\end{abstract}
\section{Introduction}
With the advancement in computer vision technology, we are witnessing an explosive surge of visual data.
Together, research in vision--language understanding has progressed significantly in the past decade,
challenging a wide variety multimodal application tasks~\cite{Wang2021SimVLMSV, Radford2021LearningTV,jia2021scaling,Alayrac2022FlamingoAV,li2023blip}, such as image captioning, visual question answering and multimodal retrieval.
With the continuing improvement in computation, researchers have extended conventional image--text models (ITMs) to 
video--text ones, mainly by substituting image encoders with 
their video counterparts~\cite{yang2021just,yang2022learning,zellers2021merlot,fu2021violet}.
This learning paradigm achieves decent performance on numerous video--text tasks, as it incorporates temporal features into modeling.
Nevertheless, 3D convolution, the core technique adopted in these video--text pretrained models, demands tremendous computational power in terms of both time and memory, limiting models' deployment on consumer-level devices. 

\begin{figure}[t!]
    \centering
    \includegraphics[scale=0.27,trim=0 0.5cm 0 0]{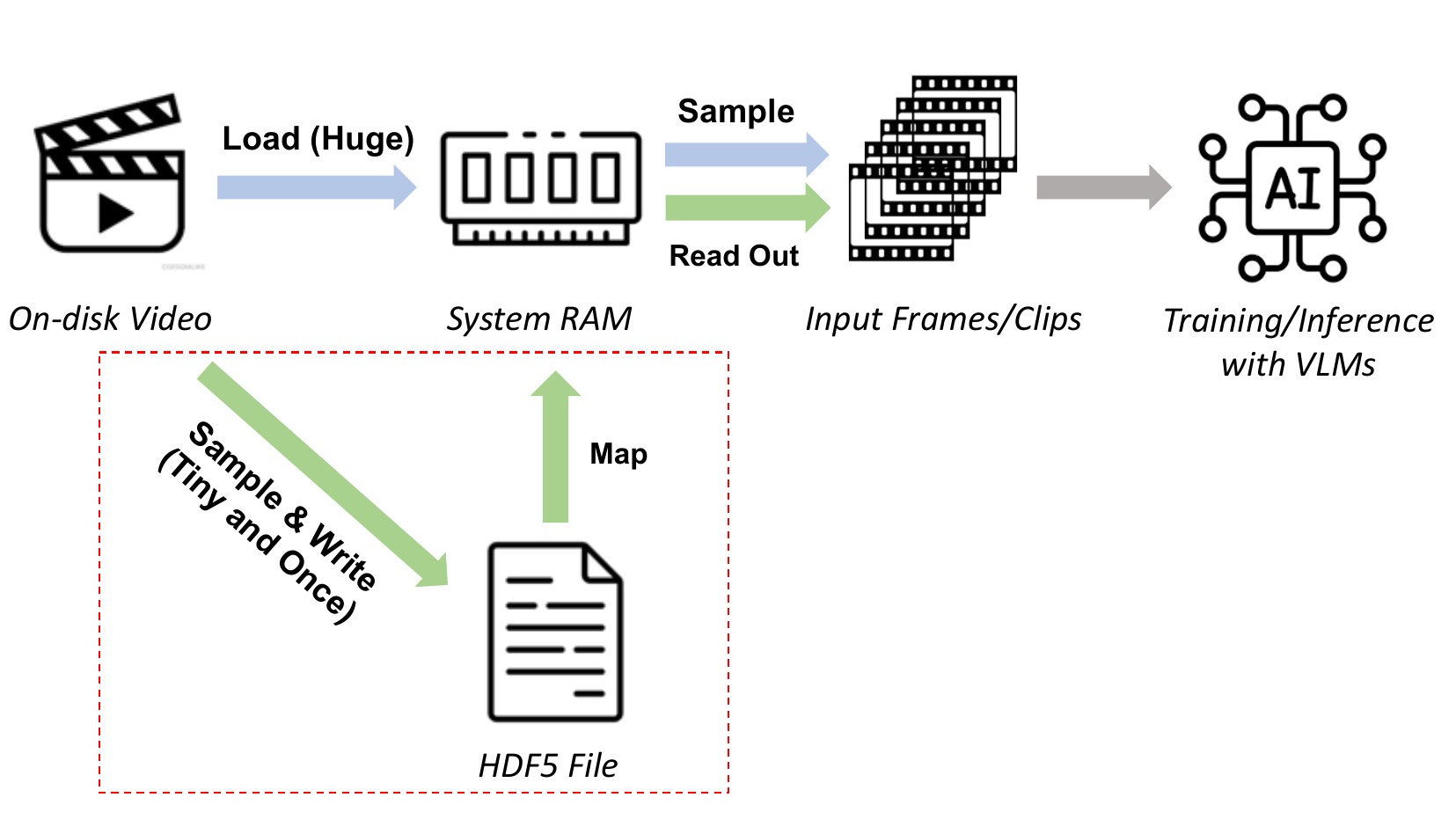}
    \caption{Comparison between conventional I/O (online sampling) and ours. The 
    blue and green arrows distinguish the dataflow between online sampling methods and ours until the end of preprocessing. The red box highlights the process we alter from conventional routines.}
    \label{fig:procedure_comparison}
\end{figure}

\begin{figure*}[ht!]
    \centering
    \def\c{0.3}
    \def\h{0.6}
    \begin{subfigure}[b]{0.49\textwidth}
        \includegraphics[scale=\c, trim={0cm \h cm 0cm \h cm}]{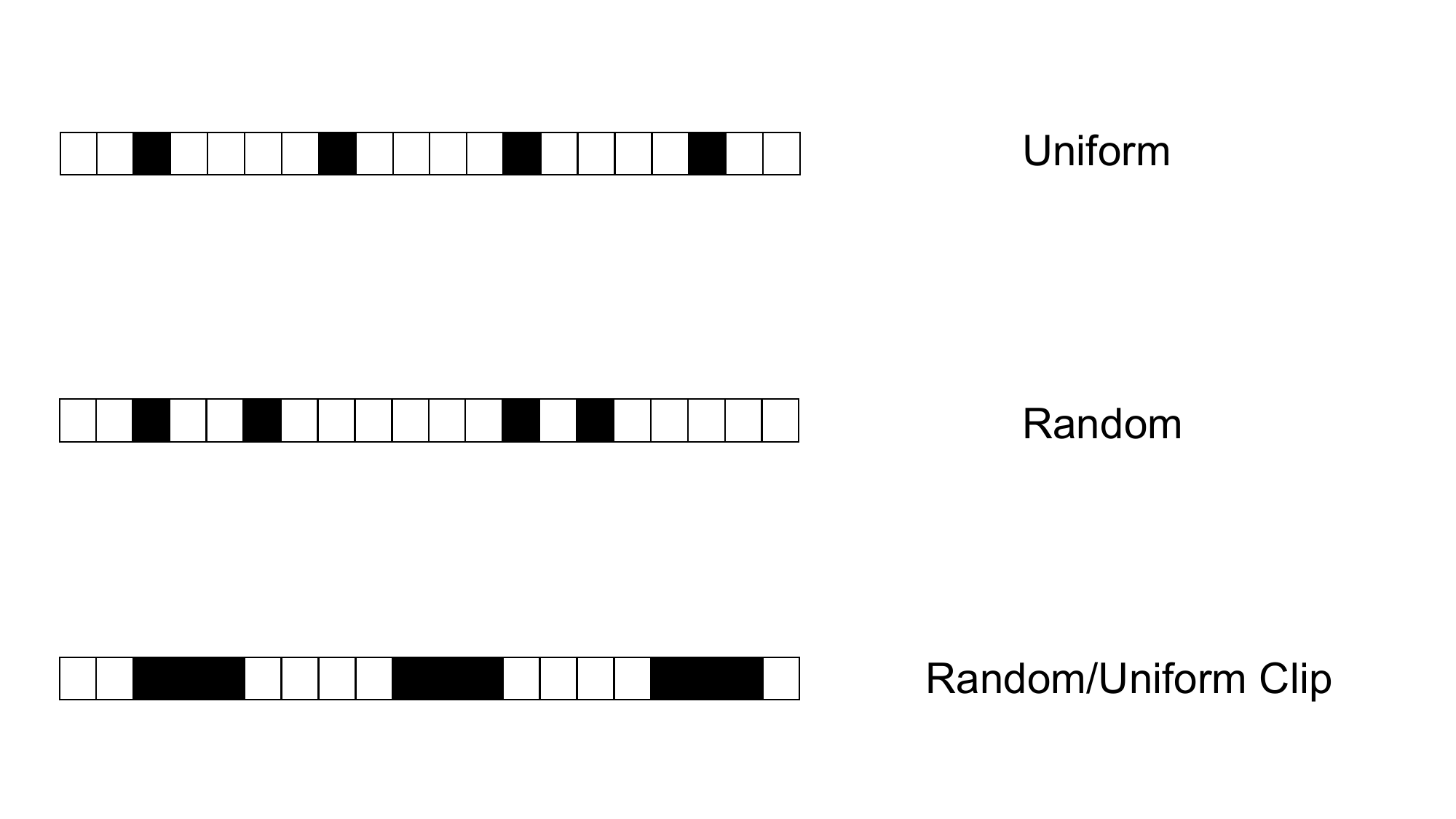}
        \caption{Heuristic Sampling}
        \label{fig:heuristic}
    \end{subfigure}
    \hfill
    \begin{subfigure}[b]{0.38\textwidth}
        \includegraphics[scale=\c, trim={0cm \h cm 1.5cm \h cm}]{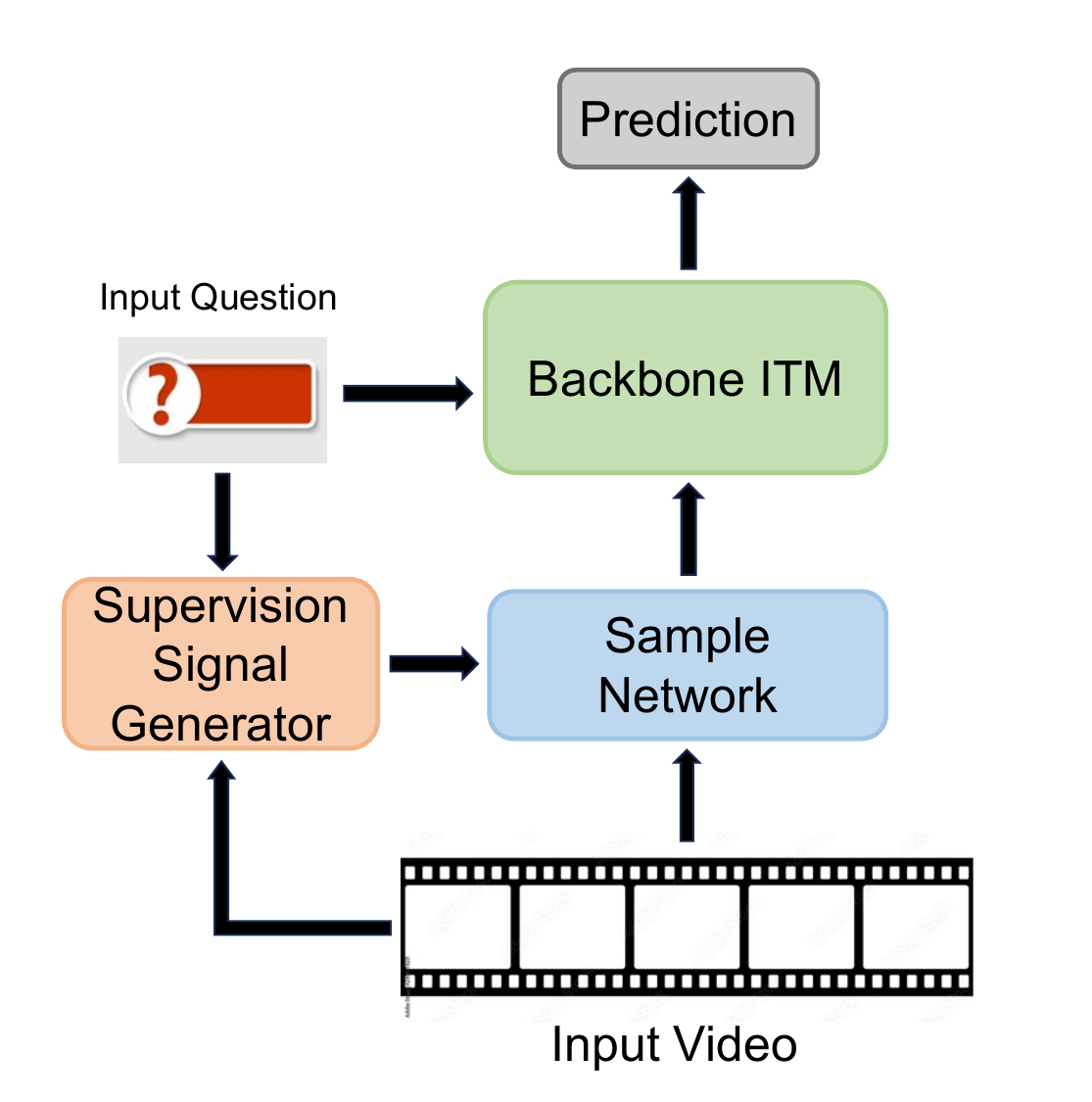}
        \caption{Learning Based Sampling}
        \label{fig:learn}
    \end{subfigure}
    \caption{Existing sample strategies for video--question answering tasks. In heuristic sampling, the black boxes indicate selected frames.}
    \label{fig:sample_strategies}
\end{figure*}

A straightforward solution to reduce overhead is to extract solely those~\textit{keyframes} that describe the main content or are related to the task from a given video, so that image--text models can preprocess them~\cite{rasheed2022fine, wang2022git, li2023blip}. 
Contemporary auto-regressive ITMs manage to adapt themselves to video--text tasks with a few frames sampled from those videos and yield promising results~\cite{rasheed2022fine, wang2022git}. 
In this family of approaches, image frames or clips (consecutive frames, as shown in~\cref{fig:heuristic}) are sampled from raw videos, cut into patches, and then encoded through a visual encoder (e.g., ResNet~\cite{he2016deep} and ViT~\cite{dosovitskiy2020image}).
X-CLIP~\cite{ni2022expanding}~further inserts cross-frame communication modules to construct connections across timestamps. 

\begin{figure}[!ht]
    \centering
    \includegraphics[width=\columnwidth, trim=2.2cm 1.5cm 0 0]{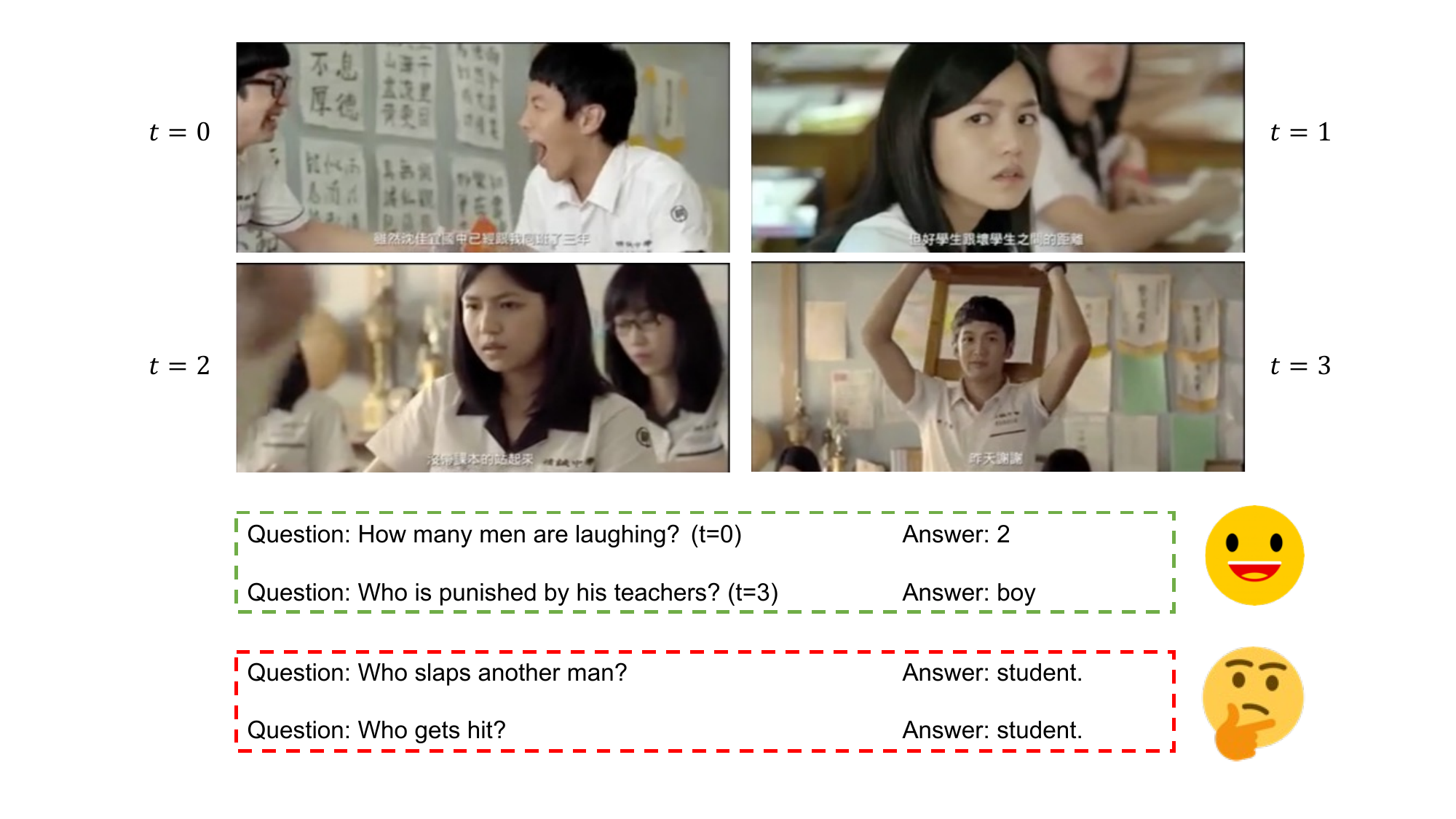}
    \caption{Randomly sampled video frames from the msrvtt-qa dataset
    and two questions. The bracketed timestamps indicate cues for corresponding answers from the video. The QA pair in the red box cannot be grounded from the four sampled frames.} 
    \label{fig:teaser}
\end{figure}

Despite performing well, we observe that the sampling strategies employed in these models are simplistic: they are blind to the video and question and only base on statistical probability distributions (\cref{fig:heuristic}).
These data-agnostic approaches inevitably limit the performance when finetuning and inferring on these ITMs, since they may cause keyframe 
omission (\cref{fig:teaser}).

On the other hand, recent works~\cite{li2022equivariant,li2022invariant,wei2023visual} introduce learning-based sampling methods.
Assisted by the Gumbel-Softmax trick~\cite{jang2016categorical}, they build a parametric sampling  network and concatenate that to the backbone. Then, as an auxiliary module, the parametric sampling strategy is jointly optimized with the main video--QA task. 
Although these frameworks gain competitive performance, they have the following drawbacks. 
First, they sacrifice efficiency owing to the additional overhead and the slow convergence speed caused by the devised sampling network, compared to direct few-frame fine-tuning on ITMs (from less than 10 epochs to more than 50 epochs)~\cite{li2022invariant,wei2023visual}. 
Secondly, it also undermines flexibility---the intervention touches the preprocessing stage in these works~\cite{li2022invariant,wei2023visual}. They encode the pre-sampled clips with customized pretrained video encoders, like 3D ResNet101~\cite{hara3dcnns} or CLIP~\cite{Radford2021LearningTV}, leading to incompatibility with ITMs which only accept raw images as input.
Additionally, the sampling network must be optimized along with the backbones on such clip features, which deters them from being directly applied to ITMs. 

To address these issues, we first explore the correlation between model's performance and the frames output from captioning-based samplers.
Specifically, we propose a learning-free sampling method, dubbed \textit{\MIF}~(MIF), which we show is an simplified and unification of previous V(isual) Q(uestion)-aware methods.
It utilizes lightweight pretrained models to annotate frames and grades each of them with a caption--question score. The selected frames are those with highest scores, or the best captions that \textit{imply} the answer.
Then, we conclude from empirical studies on MIF that capturing the most question-related frames is not a prerequisite for better accuracy. 
Based on our analysis, we hypothesize that question-aware sampling is not necessary and propose another self-adaptive sampling strategy---\textit{\MDF}~(MDF).
The underlying logic is to diversify the input frames to minimize the \textit{dominant} scenes in that video, because most of the answers can be answered from \textit{static scenes} instead of \textit{dynamic segments}.
To this end, we first define a goal function that measures the dynamics in videos whose input is the visual feature encoded by the backbone model's inherent image encoder. Then we devise a search algorithm to quickly locate the 
frames where features move slowest in that video. 
Since question content no longer participates in the sampling process, MDF is a V-aware Q-agnostic method.
In implementation, both MIF and MDF are executed in an offline fashion~\Cref{fig:procedure_comparison}, enhancing the training efficiency compared to those online sampling algorithms.
We further conduct experiments on three ITMs (CLIP~\cite{Radford2021LearningTV}, GIT~\cite{wang2022git} and All-in-one~\cite{wang2022allinone}) using four widely tested video QA datasets. 
The results show that both methods are feasible solutions towards Video--QA tasks on ITMs, among which MDF can provide better efficiency, and indirectly substantiating the correctness of our hypothesis.

The contributions in our paper are as follows:
\begin{itemize}
    \item We propose MIF, an offline question-aware sampling method for video question answering, which leverages two backbone models as captioner and scorer respectively.
    \item Based on the analysis of the MIF experimental results, we hypothesize that question-aware is redundant and propose a more efficient question-agnostic sampling method, MDF.
    \item We conduct comprehensive evaluation on a large variety of datasets and models. MDF yields competitive results with MIF, and both methods exceeds strong baselines, which also substantiates our hypothesis.
\end{itemize}

\section{Related Work}
\subsection{Visual Language Models}
Since the remarkable success of vision language models (VLMs) like  CLIP~\cite{Radford2021LearningTV}~and ALIGN~\cite{jia2021scaling}~in the field of zero-shot multimodal learning, there is a growing trend in training large VLMs through minimizing image--text contrastive loss~\cite{Li2020OscarOA,Kim2021ViLTVT,Zhang2021VinVLRV,Yu2022CoCaCC} to achieve cross-modality semantic alignment.
Early VLMs for multi-task purposes frequently adopt a bi-encoder architecture~\cite{Radford2021LearningTV,Li2021AlignBF,Li2022BLIPBL}, where visual and textual modality are separately encoded in their individual encoders and finally combined to complete downstream tasks. 
Recent achievements resort to the more efficient GPT-style~\cite{Brown2020LanguageMA} architecture, which takes the output sequences from visual encoders as the visual prefixes and jointly tunes the decoder and visual encoder~\cite{Tsimpoukelli2021MultimodalFL,Alayrac2022FlamingoAV, li2023blip}. 
When confronted with video data, a common practice~\cite{Seo2020LookBY,yang2021just} replaces image encoders in these ITMs with video encoders that can capture temporal correlations, like S3D~\cite{Xie2017RethinkingSF} and video Swin-Transformer~\cite{Liu2021VideoST}.

\subsection{Sampling Techniques in Video Question--Answering Tasks}
To apply ITMs on video understanding tasks, 
sampling is demanded to convert streaming data into discrete frames.
Most of current sampling algorithms are online algorithms, i.e., sampling happens after loading the streaming-in video data into the memory.  
The heuristic sampling methods (\cref{fig:heuristic}) are prevalent in default ITM implementations~\cite{Lei2021LessIM,fu2021violet,wang2022git,wang2022allinone}, since these algorithms are learning-free and convenient to adjust.
However, \citet{buch2022revisiting}~points out that for most video understanding tasks, understanding of event temporality is often not necessary to
achieve strong or state-of-the-art performance.
Therefore, recent works turns to integrate the sampling module into the entire learning frameworks.
As shown in~\cref{fig:learn}, this kind of architectures usually has a parameterized sampler, 
which is trained with pseudo labels generated from a question-guided indices generator and then jointly optimized with the predictions of the main task~\cite{li2022equivariant,li2022invariant,wei2023visual}.
Based on the causal theory~\cite{pearl2016causal},~\citet{li2022equivariant} separate the clips into causal and complement ones;while~\citet{li2022invariant} and~\citet{wei2023visual}~consider invariant/transient and positive/negative scenes.
Distinct to these online sampling algorithms, our proposed methods are totally offline and learning-free, but sufficiently utilizes the inherent knowledge learned by these ITMs during pretraining. 
Finally, the sampled frames are saved into HDF5 files for fast loading during fine-tuning, which greatly cut off the training time.

\section{Method}
In this section, we first briefly recap the background of the video-QA task on ITMs. 
Then we introduce the \textit{\MIF}~(MIF), a generalization to previous question--aware sampling approaches. We report some primary results and describe our key findings to the statistics. 
Finally, based on these discoveries we introduce the more efficient~\textit{\MDF}~(MDF).

\subsection{Problem Definition}
Given a short video $V=\{v_1,v_2,...,v_T\}$ of $T$ frames and a literal question $Q=\{q_1,q_2,...,q_l\}$ of $l$ tokens, an ITM $\mathcal{M}$ is expected to generate an answer $\hat{A}=\{\hat{a}_i\}_{i=1}^n$ (generative setting, $n \geq 1$) or the answer index (multiple choice setting, $n=1$) to match a reference answer which serves as a valid response to the given question.
\begin{equation}
    \hat{A} = \mathcal{M}(V', Q) 
\end{equation}
where $V'\subset V$ is the set of sampled frames.

In evaluation, we use item-wise accuracy as the performance metric, defined as:
\begin{equation}
    acc = \frac{1}{\vert \mathbf{Q} \vert} \sum_{i=1}^{\vert\mathbf{Q}\vert}\mathbf{1}(\hat{A}_i=A_i)
\end{equation}
where $\mathbf{Q}$ is the entire set of questions in the dataset, $\mathbf{1}(\cdot)$ is the indicator function that equals 1 only if the expression is true. 
The predictions can be either generated through direct generation (generative setting) or classification (multiple choice setting). See~\cref{sec:eval_metric}~for more details.

\subsection{Most Implied Frames (MIF)}
\label{sec:mif}
\begin{figure}[ht!]
    \centering
    \includegraphics[scale=0.23, trim=1.5cm 1.5cm 0 0]{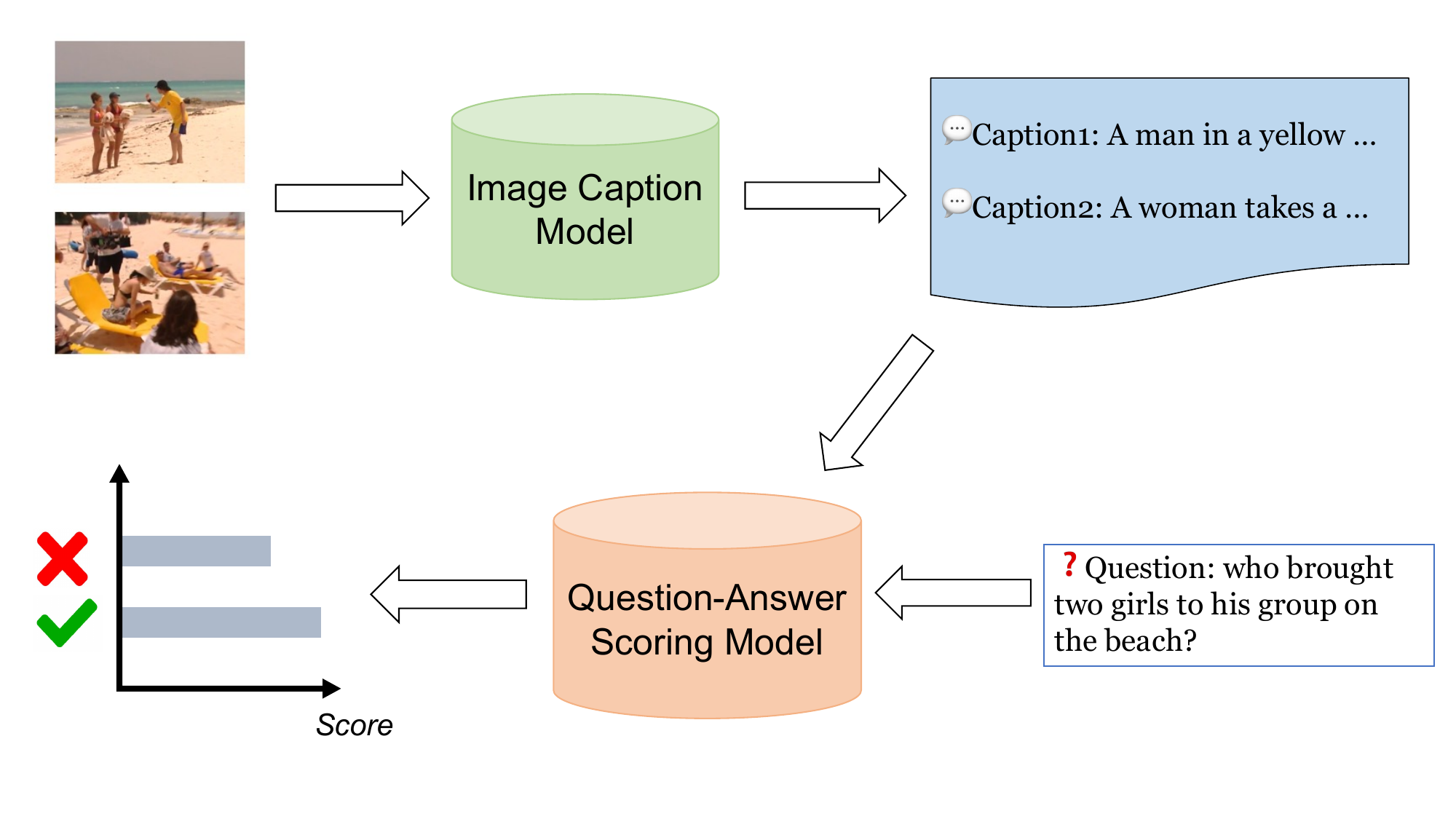}
    \caption{MIF workflow. Here we just show an example of how it selects one frame out of two frames.}
    \label{fig:mif}
\end{figure}

MIF uses a caption model $\mathcal{M}_c$ and a set of grading models $\mathcal{M}_g$ to select the best frame candidates, as illustrated in~\cref{fig:mif}. Given a question, MIF could also be termed ``cue frame retrieval''.
Before starting the process, following previous work~\cite{buch2022revisiting,li2022invariant}, we reduce the computational cost by uniformly sampling $T'$~($T' << T$)~frames from the original video, with indices as $\{t_1,t_2,...,t_{T'}\}\subset\{1,2,...,T\}$. 
The caption model $\mathcal{M}_c$ takes all downsampled frames as input and generates a description $C$. 
Then $\mathcal{M}_g$ computes the matching score $s$ between the question $Q$ and the generated description ($s=\mathcal{M}_g(Q,C)$). 
We presume that the matching score $s$ indicates the possibility that each frame can serve as a cue to answer the given question. 
Therefore, we rank all frames by score, selecting the highest $N$ frames as the sample (indicated by indices):
\begin{equation}
    i_1,i_2,...,i_N = \argtopk_t(\{s_{t_1},s_{t_2},...,s_{t_{T'}}\},N) 
\end{equation}
where $s_t$ is the matching score for frame $v_t$. 
Notably, MIF is a QA--aware algorithm. For questions posted under the same video, MIF usually generates different sets of sampling results.

\subsection{Primary Results on MIF}
The main results by MIF can be found in~\cref{tab:res_clip},~\cref{tab:res_other}~and~\cref{tab:aio}. All experiments leverage the base version of GIT (consistent with target model) to generate captions and BERT\footnote{https://huggingface.co/iarfmoose/bert-base-cased-qa-evaluator}~finetuned on many prevalent textual question answering datasets (SQuAD~\cite{rajpurkar2018know}, RACE~\cite{lai2017race}, CoQA~\cite{reddy2019coqa}~and~MSMARCO~\cite{nguyen2016ms}) as the grader to calculate question--caption correlation score. 
The increment of accuracy is significant on all backbone models and datasets compared to state-of-the-art baselines, showing that MIF is a promising solution when performing video understanding tasks on ITMs. 

Upon the decent performance, we are curious about the correlation between accuracy and captioner/grader model sizes in MIF---for which we form our first research question below.
\paragraph{RQ1:} Are stronger captioning or scoring models bound to bring better results?

To provide a potential response, we systematically study MIF by testing frames picked via two general types of samplers on GIT-Base: i) two separate models for captioning and grading; ii) BLIP-2 pretrained on QVHighlights~\cite{lei2021detecting}~as a unified model for question-aware key-frame extraction~\cite{yu2024self}.
\begin{table}[ht]
    \small
    \centering
    \resizebox{0.9\linewidth}{!}{
        \begin{tabular}{c|c|cc}
        \toprule
             $\mathcal{M}_c$ & $\mathcal{M}_g$ & MSVD & MSRVTT  \\
            \midrule
            \multicolumn{4}{l}{\textit{\textbf{Separate Model}}} \\
            \midrule
            GIT-S & BERT-S & 46.5 & 42.3  \\
            GIT-B & BERT-B & 46.7 & 42.4  \\
            GIT-L & BERT-L & 46.9 & 42.1  \\
            \midrule
            \multicolumn{4}{l}{\textit{\textbf{Unified Model}}} \\
            \midrule
            \multicolumn{2}{c|}{BLIP2-T5-XL} & 46.6 & 42.0 \\
            \multicolumn{2}{c|}{BLIP2-T5-XXL} & 46.2 & 42.2 \\
        \bottomrule
        \end{tabular}
    }
    \caption{Results of MSVD-QA and MSRVTT-QA on GIT using frames sampled from different captioner-grader combinations. The number of input frames are fixed at 6. ``GIT-B" and ``Bert-B" is the default implementation in later sections.
    }
    \label{tab:mif_sizes}
\end{table}

Among these results, we find that there is no significant correlation between the size of caption-grading system and the accuracy of Video--QA task, though larger models may produce more informative and accurate captions and scores overall. 
Now that question-guided sampler has reached its upper bound, 
we make a bold hypothesis: \\ 

\noindent 
\textbf{Hypothesis:} \textit{Question-agnostic sampling methods can perform as well as question-aware ones.}

\paragraph{RQ2:} Can we design a question-agnostic sampler?
To provide a possible solution, we propose another method, \textit{\MDF}~(MDF), in the following section, powered by the inherent vision-encoder of ITMs.

\subsection{Most Dominant Frames (MDF)}
It has been pointed out in early video sampling works~\cite{shahraray1995scene,nam1999video}~that the sampling rate in each temporal region should be proportional to the object motion speed. 
Besides, because the frame lengths are usually fixed in ITMs (3 or 6 in our experiments), if the sampled frames are temporally closed, at a large chance they will share analogous contents and some key frames may be missing.  

\begin{figure}
    \centering
    \includegraphics[scale=0.43, trim=0 2cm 0 1cm]{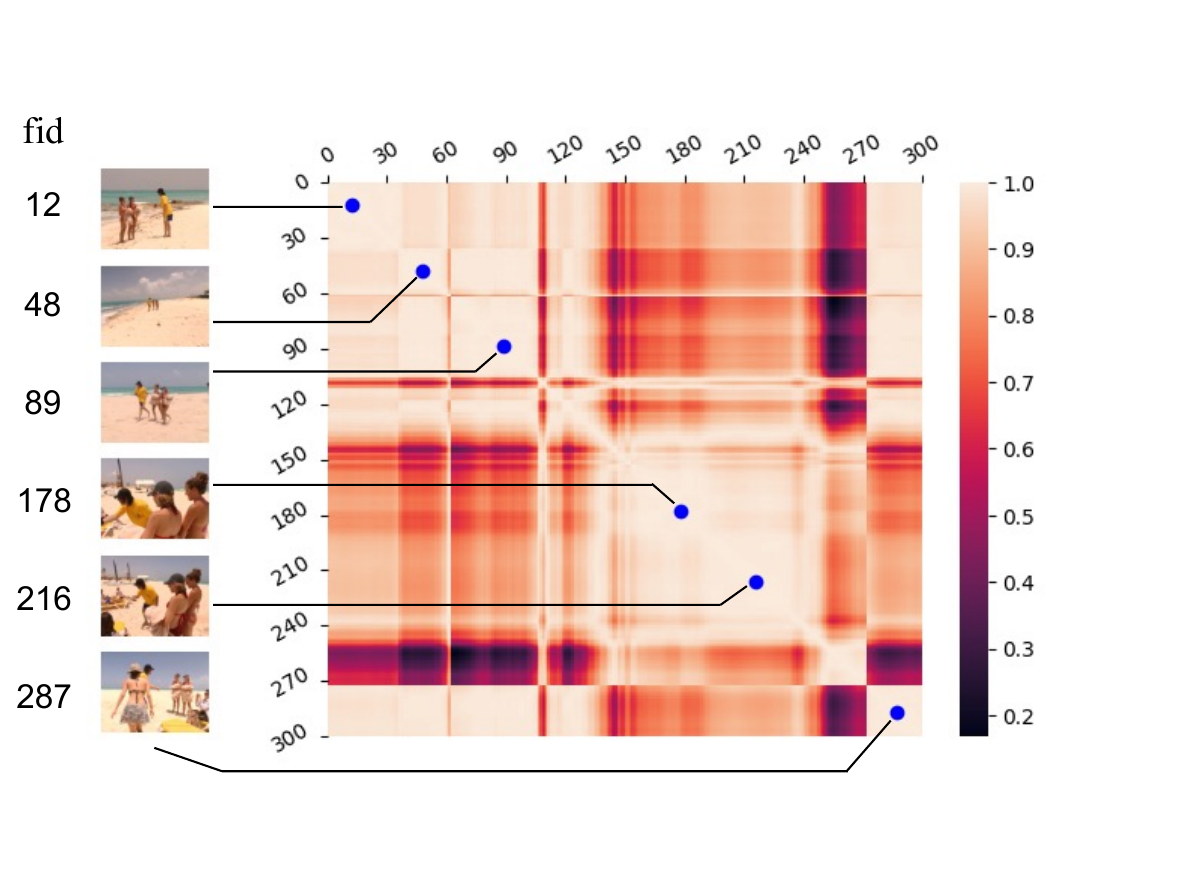}
    \caption{Sample MDF processing (6 frames). 
    The heatmap visualizes the calculated frame similarity matrix as the cosine value between pairs of frame vectors. 
    The entry at $i^{th}$ row $j^{th}$ column represents the similarity between frames $i$ and $j$. 
    Blue points indicate the frames eventually extracted.}
    \label{fig:mdf}
\end{figure}

To this end, we construct our solution based on the ITM's cognition towards the frames from its own vision module.
The first intuition comes from the theory and experience of representation learning from large pretrained models~\cite{bengio2013representation,devlin2018bert,dosovitskiy2020image}, where learned representation output from well-tuned large models embed meaningful semantic information. 
We harness the inherent vision encoder of the VLM (if it has one) to acquire visual embeddings $E=\{e_1,e_2,...,e_T\}$.
To quantify the invariance in each frame, we define the following metric $dom(t)$ (the abbreviation of dominant) for frame $v_t$ at timestamp $t$.
\begin{equation}
    dom(t) = \sum_{t'=t-W}^{t+W}\mathbf{sim}(e_t,e_t')
    \label{eq:dom}
\end{equation}
The problem then can be formulated as seeking $N$ local minima of $dom(t)$ with respect to time $\tau=\{t_1,t_2,...,t_N\} \subset \{1,2,...,T\}$, subject to $\vert\tau_i-\tau_{i+1}\vert \geq W$.

\begin{algorithm}
\small
\SetKwFunction{ret}{return}
\SetKwFunction{argtop}{argtop}
\KwIn{Video frames $V=\{v_1,v_2,...,v_T\}$, vision model $\mathcal{M}$, width-adjusting rate $\lambda$}
\KwOut{Visual prefix $F=\{f_1,f_2,...,f_N\}$}
Encode frames using the vision model $E=\mathcal{M}(V)=\{e_1,e_2,...,e_T\}$ \\
Compute $dom$ score for all frames and set $W$, according to ~Eq. \ref{eq:dom}~and~Eq. \ref{eq:W}. \\
\textbf{Init} $F=\{f_{\arg \max_t dom(t)}\}$, index set $I=\{0,1,...,i-W, i+W,...,T \}$ \\
\While{$\vert F \vert < N$ and $I \neq \oslash$}{
    $t' \leftarrow \arg \max_t dom(t)$\ \\
    $F \leftarrow  F \cup \{{f_{t'}}\}$ \\
    $I \leftarrow I \setminus \{t''\}_{t''-t'<W}$ \\
}
\uIf{ $\vert F \vert < N$ }{
    $\tau \leftarrow \argtop_{N}(\{dom(t)\}_{t \in T})$ \\
    \ret $F \cup \{f_t'\}_{t'\in \tau}$ \\ 
}\Else{
    \ret $F$ \\
}
\caption{Most Dominant Frames (MDF)}\label{alg:mdf}
\end{algorithm}

The details of the algorithm is given in~\cref{alg:mdf}.
Considering the disparity in the lengths of videos, instead of keeping a constant $W$, we set $W$ automatically in an self-adaptive way:
\begin{equation}
    W_V = L_V / (\lambda \cdot N)
    \label{eq:W}
\end{equation}
where $L_V$ is the length of video $V$ in terms of frame numbers, $\lambda$ is the constant width-adjusting rate that controls the scope to search in every steps.
\cref{fig:mdf} visualizes an example of searching results on the similarity map.

\section{Experiments}
\textbf{Datasets}. To evaluate our proposed methods, we conduct extensive experiments on the following 4 frequently tested datasets:
\paragraph{MSVD-QA and MSRVTT-QA.}
These two datasets~\cite{xu2016msr} are adapted from corresponding video captioning datasets---Microsoft Research Video Description Corpus~\cite{Chen2011CollectingHP} and Microsoft Research Video to Text~\cite{Xu2016MSRVTTAL}. 
Both datasets provide same five types of questions---\textit{what, where, who, when, how}.
The answers to the questions are all single words.

\paragraph{TGIF-QA.} The TGIF-QA~\cite{jang-IJCV-2019} dataset contains 165K QA pairs for the animated GIFs from the TGIF dataset~\cite{li2016tgif}. Its question--answer pairs are annotated via crowdsourcing with a carefully designed user interface to ensure quality.
TGIF-QA offers three question types: frame, transition, and (repetition) count. We folllow previous common benchmarking work~\cite{fu2021violet,wang2022git,xiao2022video} and test only on the frame-QA task. 

\paragraph{NExT--QA.}
The NExT-QA dataset~\cite{xiao2022video} targets at reasoning from causal and temporal relationships between actions. 
There are three question types in NExT--QA: descriptive, temporal and causal reasoning, which respectively targets at evaluating model's different aspects of capability.
There are two versions for the composition of questions and answers: open-ended and multiple choice (MC). 
We test our methods on the MC setting following the most common practice. 

\subsection{Backbone Models}
\paragraph{CLIP}
CLIP~\cite{rasheed2022fine} is the first ITM that focuses on zero-shot transfer onto diverse multimodal downstream tasks. 
It is composed of two modality-specific encoders to process input modality signals separately. 
In our experiments, we also modify its structure by adding a single-layer transformer decoder on the top of the two encoders (dubbed ``CLIP-dec'' but we still use "CLIP" to denote it for simplicity). 
We decode for only one step to get the answer, not alike other generative ITMs that predict the whole sequence containing both the question and answer words.

\paragraph{GIT}~\cite{wang2022git} is one of the state-of-the-art ITMs for video question answering tasks, released by Microsoft Research.
It adopts ViT-B-16~\cite{Radford2021LearningTV} as its visual encoder and a GPT-style decoder that receives both the encoded image patches (as visual prefix) and textual embeddings to generate the output text.
Currently the GIT family consists of four versions\footnote{GIT-Base, GIT-Large, GIT and GIT2, as of July 2023}.
In our experiments, we tune GIT-Base on these three datasets (denoted as GIT in later context for simplicity).

\paragraph{All-in-one (AIO)}~\cite{wang2022allinone} is another family of ITMs which follows the philosophy of \textit{learning-by-fusion}.
The model is composed of stacked multimodal attention layers called a unified transformer that takes concatenated video--text input as the basic fusion modules.
Similar to the previous two ITMs, it can adapted to employ output embeddings to solve many downstream video--language tasks.
Particularly, we use All-in-one(-Base) in all our experiments.

In what follows, by default ``CLIP'' and ``AI'' respectively denote CLIP-ViT-base-patch16\footnote{https://huggingface.co/openai/clip-vit-base-patch16} with a decoder and All-in-one-Base\footnote{https://github.com/showlab/all-in-one}.
For GIT-related models, we follow~\cite{wang2022git} to finetune the pretrained GIT-Base\footnote{https://huggingface.co/microsoft/git-base}~on four datasets).

\subsection{Baselines}
\paragraph{Direct Finetuning} We first consider directly finetuning each backbone model, which can be categorized into online learning-free sampling. 
Since the exact sampling strategy adopted by GIT is unknown, we examine the results using uniform sampling and find that they are closed to the reported numbers on three datasets. Hence, we treat uniform sampling as baseline for GIT and CLIP-series (because there is not open-sourced implementation provided for CLIP on these datasets as well). 
As AIO provides public code, inclusive of sampling strategy, we report such baselines results direct using their code (inclusive of their hyperparameter settings) for both training and testing.

\paragraph{Learning-based Sampler}
We compare with two advanced learning-based samplers, IGV~\cite{li2022invariant}~and~VCSR~\cite{wei2023visual}. Both methods construct two or more complement segment groups with 
contrastive
property and jointly optimize the main network and sampler by minimizing 
auxiliary losses.
In original implementation, both IGV and VCSR sample much more frames than the default input lengths of backbone ITMs ($\vert V \vert = 16$ in IGV and $\vert V \vert = $frames/clip$\times$clip $ = 6\times 4=24$ in VCSR) to the same value ($1 \times 3$ for VCSR).
Because enlarging the input size improves accuracy (see~\Cref{sec:frame_num}), for fair comparison we reset the sampling size when implementing the two methods on each backbone model.

\subsection{Implementation Details}
The details of MIF have been introduced in~\Cref{sec:mif}.
In MDF, we use each model's inherent vision encoder to encode the sampled frames, and then calculate the cosine values between these vectors as the measure of frame similarity.
A special case is that AIO does not have an independent visual encoder. 
Hence, we use ViT-B-16 (the same visual encoder as CLIP and GIT) as the ``pseudo visual encoder'', and following the same procedure to obtain the sampled frames in each video.

\begin{table}[ht]
    \centering
    \resizebox{\linewidth}{!}{
    \begin{tabular}{l|ccc}
    \toprule
        Model & MSVD & MSRVTT & TGIF  \\
    \midrule
        Base~{\cite{Radford2021LearningTV}} & 33.8 & 33.7 & 59.9 \\
        IGV {\cite{li2022invariant}} & 34.8 & 34.1 & 61.9 \\
        VCSR {\cite{wei2023visual}} & 34.6 & 34.5 & 61.6 \\ 
    \midrule
        MIF (Ours) & 35.0  & \textbf{35.4}  & 62.5  \\
        MDF (Ours) & \textbf{35.1}  & 35.2  & \textbf{63.2}  \\
    \bottomrule
    \end{tabular}}
    \caption{Experimental results on CLIP ($\vert V \vert = 3$) backbone and three datasets. 
    }
    \label{tab:res_clip}
\end{table}

\subsection{Results}
\paragraph{Results on CLIP}
~\cref{tab:res_clip} shows the results over the three datasets. 
Both MIF and MDF acquire achieves significant improvement over original CLIP implementations (1.2$\sim$3.3\%) and baselines that incorporate learning-based sampling methods. 
However, the performance gap between the sampling strategies is insignificant on both MSVD-QA and MSRVTT-QA, indicating that question awareness is unnecessary for performance.

\begin{table}[ht]
    \centering
    \resizebox{\linewidth}{!}{
    \begin{tabular}{l|ccc}
    \toprule
        Model & MSVD & MSRVTT & TGIF  \\
    \midrule
         \multicolumn{4}{l}{\textit{\textbf{GIT Backbone}}} \\
    \midrule
        Base~\cite{wang2022git} & 52.2 & 41.1 & 67.5 \\
        IGV~\cite{li2022invariant} & 53.2 & 41.5 & 68.1 \\
        VCSR~\cite{wei2023visual} & 52.7 & 41.6 & 68.6 \\
    \midrule
        MIF & 54.5 & \textbf{42.3} & 69.9 \\
        MDF & \textbf{55.3} & 42.0  & \textbf{70.0} \\
    \midrule
        \multicolumn{4}{l}{\textbf{\textit{AIO Backbone}}} \\
    \midrule
        Base~\cite{wang2022allinone} & 46.1 & 42.7 & 64.0 \\
        IGV~\cite{li2022invariant} & 46.3 & 43.3 & 64.7 \\
        VCSR~\cite{wei2023visual} & 46.4 & 43.0 & 64.5 \\
    \midrule
        MIF & 46.7 & \textbf{44.0} & 65.9 \\
        MDF & \textbf{46.9} & 43.8  & \textbf{66.2}  \\
    \bottomrule
    \end{tabular}}
    \caption{Test set results on MSVD, MSRVTT and TGIF. Best scores are bolded. }
    \label{tab:res_other}
\end{table}
\begin{table}[ht]
    \centering
    \resizebox{0.8\linewidth}{!}{
    \begin{tabular}{l|cc}
    \toprule
        Model & Val & Test  \\
    \midrule
        Base~{\cite{wang2022allinone}} & 47.1 & 45.9 \\
        IGV {\cite{li2022invariant}} & 48.3 & 47.1 \\
        VCSR {\cite{wei2023visual}} & 48.0 & 47.4 \\ 
    \midrule
        MIF (Ours) & 48.5 & \textbf{48.2}  \\
        MDF (Ours) & \textbf{48.8} & 48.0 \\
    \bottomrule
    \end{tabular}
    }
    \caption{Results on validation and test of the multi-choice NExT-QA dataset (5-choices per question).}
    \label{tab:aio}
\end{table}

\paragraph{Results on GIT and AIO.} 
\Cref{tab:res_other}~and~\Cref{tab:aio}~display the results of GIT and AIO on four datasets.
There are three key points to worth concerning.
Firstly, compared to the original implementation results, both MIF and MDF can enhance the accuracy on all four datasets regardless of model architectures. 
This appearance matches the trend on CLIP, which demonstrates our proposed methods are broadly applicable to diverse datasets and models. 
Secondly, the increment in accuracy is higher on models with more sampled frames (6 in GIT and 3 in AIO), which implies that our proposed methods are possibly more effective when the input frames is longer.
Lastly, we notice that the improvement on TGIF-Frame by MIF and MDF over VCSR is more drastic than the other two datasets. 
This outcome is somehow counter-intuitive since videos in TGIF-Frame are much shorter with fewer chance in switching scenes---by intuition the dataset should be more insensitive to the sampling variants.

\section{Analysis}

\subsection{Impact of Input Frame Length}
\label{sec:frame_num}
\begin{figure}[t!]
    \begin{subfigure}[t]{\linewidth}
    \centering
    \includegraphics[scale=0.5,trim=2cm 0cm 0 0]{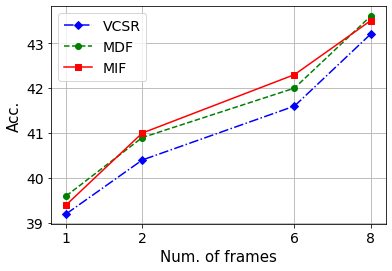}
    \caption{}
    \label{fig:perf_nofr}
    \end{subfigure}
    \begin{subfigure}[t]{\linewidth}
    \centering
    \includegraphics[scale=0.5,trim=0.4cm 0cm 0 0]{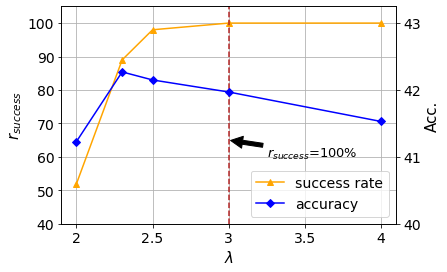}
    \caption{}
    \label{fig:perf_intv}
    \end{subfigure}
    \caption{Performance compared to VCSR~\cite{wei2023visual}~under (a) different input lengths of frames in both MDF and MIF (b) varied separation factor $\lambda$ in MDF on the MSRVTT-QA dataset by GIT.}
\end{figure}

Recall that we fix all baselines' input frame lengths in all experiments.
However, intuitively the number (length) of input frames should be regarded as a potential factor to the accuracy, since increasing the input frames equals to exposing larger amount of training data to the model.
To see how this factor affects backbone models' performance and whether our proposed sampling methods can consistently enhance the accuracy when sampling more or fewer frames, we continue to fine-tune GIT on the MSRVTT--QA dataset with distinct frame lengths.
The results of this set of experiments are plotted in~\Cref{fig:perf_nofr}.
From the figure we firstly discover that as expected, after increasing the number of input frames, the accuracy scores become higher.
Moreover, the accuracy of the proposed two sampling strategies MDF and MIF consistently surpasses the VCSR baseline, indicating that they can really locate those key frames in videos even after changing the input length.

\subsection{Auto-generated Captions in MIF}
In MIF, we invoke a captioning model and anticipate it to provide precise and informative annotations to each frame.
Since intuitively, the question--answering matching judgement model can not probably differentiate nuance in two sentences if their pattern looks quite similar.
However, the actual results are opposite to our expectation.
Take our randomly selected video from MSVD-QA in Table~\ref{tab:mif_example}~as an example, where Q1 and Q2 denote two questions ``what does a small dog wildly play with?'' and ``what wildly plays with a ball?''.
First, it can be observed that the captions generated by the VLM looks similar to each other, in the format of `` [\textit{noun}] [\textit{verb}] [\textit{prep. phrase}]'', suggesting that the captioner model tends to generate descriptions in a nearly fixed pattern.
This outcome can be viewed as a syntactic bias during generation.
Moreover, the sentence similarity among these captions confuse the scorer model---although Q1 and Q2 describe nearly the same scenario and thus should share some cue frames, the most essential frame (the 12$th$ frame) is successfully captured for Q1 but discarded for Q2, as well as the second most important frame (the 3$rd$ frame).
Therefore, we believe that a captioning model that can provide diversified output and a robust scoring model that offers objective and fair ratings to question--answer pairs is necessary to guarantee sampling effectiveness, itself vulnerable to noise.

\begin{table}[h]
    \centering
    \resizebox{\columnwidth}{!}{
    \begin{tabular}{c|c|c|c}
    \toprule
    FID & Caption & Q1 & Q2  \\
    \hline
    1 & a puppy playing with toys. & & \\ 
    \hline
    2 & a white puppy playing with a toy. & & \\ 
    \hline
    \multirow{2}{*}{3} & \multirow{2}{*}{\shortstack{a white puppy with black eyes and \\ a blue ball.}} & \multirow{2}{*}{\cmark} & \multirow{2}{*}{} \\ 
    ~ & & \\
    \hline
    4 & a puppy that is laying down on the floor. & & \\ 
    \hline
    5 & a puppy playing with a blue ball. & & \\ 
    \hline
    6 & a puppy that was found in a house. &  & \cmark \\ 
    \hline
    7 & a puppy that is laying down on the floor. & & \\ 
    \hline
    8 & a puppy that is sitting on the floor. & & \cmark \\ 
    \hline
    9 & a puppy is sitting on the floor. & \cmark & \cmark \\  
    \hline
    10 & a white puppy sitting on a table. & & \cmark \\  
    \hline
    11 & a white puppy laying on the floor. & \cmark & \cmark \\   
    \hline
    12 & a puppy playing with a blue ball. & \cmark & \\   
    \hline
    13 & a white dog standing on top of a floor. & \cmark & \cmark \\    
    \hline
    14 & a white dog walking on the floor. & \cmark & \\    
    \hline
    15 & a small white dog playing with a ball. & & \\    
    \hline
    16 & a dog chewing on a toy in a cage. & & \\    
    \bottomrule
    \end{tabular}}
    \caption{Example frame captions and sampling results. ``\cmark''~marks frames chosen to constitute the input frame set along with the question in that column.}
    \label{tab:mif_example}
\end{table}

\subsection{Sampling Interval in MDF}
In MDF, we prevent the sampling frames from being excessively close by setting a hyperparameter $\lambda$ and thus the search interval $W=L/(\lambda\cdot N)$.
However, decreasing $\lambda$ (enlarging the interval $W$) incurs more frequent failure for MDF to sample enough frames, and in this case
some of the sampled frames may get too closed to degrade the target model's performance.
In our experiments, we surprisingly found that such situations do not always happen.
To delve into this phenomenon, we define the outcome where the collected $K$ frames satisfy the interval requirements as ``success'' and otherwise as ``failure''.
We test and plot the curve of success rate ($r_{success}=n_{success}/n_{total}$) and accuracy against $\lambda$ on three datasets produced by GIT, as shown in~\Cref{fig:perf_intv}. 
The horizontal axis denotes the hyperparameter $\lambda$ that controls the minimal sampling interval. 
The figure shows that there is a critical point that failure will never happen if continuing to increase $\lambda$---we do not know the precise value but only to mark the minimal value among these settings that we can earn 100\% success.
Moreover, there is no strong correlation between the success rate and model performance, but a minimum interval should be reached to ensure a promising performance.
The performance peak is achieved under a hybrid sampling strategy ($\lambda=2.3, r_{success}=79.1\%)$.

\section{Conclusion}
In this paper, we focus on the frame sampling issue inhering in the task of video question--answering and propose two simple and effective methods---\MIF~(MIF)~and~\MDF~(MDF).
MIF streamlines a set of sampling methods in the textual space by projecting heterogeneous inputs (question and video) to a common space through pretrained ITMs. It then identifies frames with the highest matching scores generated from a scoring model.
Based on the insights and analysis derived from MIF, we further propose~\MDF~(MDF), which exploits a more concise, self-adaptive formulation for sampling.
The success on these sampling strategies from CLIP to All-in-one demonstrates the broad applicability of our proposed methods across a spectrum of general scenarios.

\section*{Limitations}
Despite the promising results gained from our methods, on a wider horizon we still note unaddressed limitations.
First, due to the restriction of computation resource, we only evaluate our proposed methods on the video question answering task, and we do not have the opportunity to test on more emerged ITMs to further substantiate our methods' efficacy.
Secondly, we do not try MIF-style methods on large language models like GPT-4. These areas may serve as future directions. 

\bibliography{anthology,custom}
\bibliographystyle{acl_natbib}

\appendix
\section{Implementation Details}
To enforce a fair comparison, 
we run both training and testing stages for each VLM on a single NVIDIA RTX-A6000 GPU (except All-in-one because its implementation only has multi-GPU version, therefore we run it on 2 GPUs) while holding other hyperparameters and settings consistent with the default ones introduced in their original papers or codes (e.g., number of frames sampled per video, learning rate, training epoch, numerical precision in computation, etc). 
Gradient accumulation is applied to enable a large batch size ($\geq512$) required in the fine-tuning process.  
To further reduce the computational complexity, all experiments are implemented with the pytorch Automatic Mixed Precision (AMP)~\footnote{https://pytorch.org/docs/stable/amp.html}~package.
The checkpoints in our finetuning stage can all be found and downloaded from publicly available links. 

\section{Baseline Models}
We compare the results on the listed image--text pretrained models to other models in similar sizes that have
(1) an image encoder inside but experience no or a different pretraining procedure (including the pretraining task selection and design, the goal function, datasets and annotation methods, etc)~\cite{huang2020location,jiang2020divide,liu2021hair,Lei2021LessIM}.
(2) a video encoder to tune during training time or merely use feature vectors  extracted from pretrained video networks (I3D~\cite{carreira2017quo}, S3D~\cite{xie2018rethinking})~\cite{xiao2022video,zellers2021merlot,yang2021just,fu2021violet}. 
For baselines that work as our backbone network and finetuning starting point, we report our reproducing results as a more accurate benchmark, since we found many of these results are distinct from those reported in the original paper owing to the disparity in implementation environments.

Particularly, since we do not find any details introduced in the paper or official implementations online regarding the sampling strategies in GIT, 
and our implementation with uniform sampling in both training and testing can achieve comparable results as the reported ones~\cite{wang2022git} on 2 of 3 datasets, we treat this implementation as the reproduced results of GIT standalone.

\section{Evaluation Metrics}
\label{sec:eval_metric}
In all models, the sampled raw frames $V'$ are resized to match the model-acceptable scales and then normalized.
VLMs then take these frames as input and embed them into a sequence of vectors. 
Since the decoding mechanisms are different in these models, we illustrate them one by one:

In non-generative Video--LM (CLIP), the outputs from both modality encoders first pass through a transformer decoder layer and a classification layer:
\begin{equation}
    \hat{A}=f(E_v,E_q)
\end{equation}

In generative VLM (CLIP-Dec, GIT), the visual (from the visual encoder, like a prefix prepended to the text) and textual embeddings (from the embedding layer) constitute the input of the decoder.
The decoder keeps generating the whole question and answer sequence in an auto-regressive manner:
\begin{equation}
   P(Q,A|V,Q)  = \sum_{t=1}^{n+l-1}\log P(y_{t+1}|y_1,y_2,...,y_t,V)
\end{equation}

In All-in-one, the model first generates answer predictions $z_i$ for each frame.
Then, these predictions are fused together by summation to form a consensus at the video level~\cite{wang2022allinone}.
\begin{equation}
    p=\frac{1}{S}\sum_{i=1}^S z_i
\end{equation}

\section{Speedup and Overhead Analysis}
\paragraph{From video--text models to image--text ones.} 
By adopting image--text VLMs (even without HDF5 as storage), we can obtain a $2.5\sim 4\times$ acceleration during training and inference stage.
Moreover the training can be completed with a single A6000 GPU (46 GB memory) for all image--text VLMs in our experiments (for all-in-one although it runs on 2 GPUs, the total memory usage can fit to a single GPU, i.e., much less than 46 GB), while video--text VLMs listed as our baselines (e.g., MERLOT~\cite{zellers2021merlot}) consume 4 same type of GPUs with the same batch size.

\paragraph{From \textit{on-the-fly} sampling to offline sampling plus HDF5 I/O.} Conventional approaches for image--encoder based VLMs to generate input frames directly read from raw videos and then sample frames among them \textit{on-the-fly}, which consumes a large amount of storage and running time during training.
As our proposed methods are \textit{offline algorithms}, we can save all sampled frames for each video into a unified HDF5 file and meanwhile create a vid-to-id mapping file, (a.k.a. meta data) for the model to look up during its running time. 
HDF5 (Hierarchical Data Format) is a file format designed to store and organize large amounts of data by creating a set of "datasets", and to address current and anticipated requirements of modern systems.
The contents saved in an HDF5 file can be mapped to RAM for fast loading during training, which greatly reduces the time needed for model training.

As a direct comparison, in our implementation of All-in-one, a $2.5\sim 2.9 \times$ speed-up during training stage is recorded when using HDF5 to substitute original reading from video-files and then sampling \textit{on-the-fly}. 
For GIT and CLIP, this kind of comparison is infeasible since the training time can not be found neither in their papers nor replicated by our implementations (since we do not find open-sourced code for them on these video--QA datasets, the replication of their results also adopts the HDF5 I/O).

\paragraph{Removal of Redundant Sampling.}
Although the sampling process in the preprocessing stage produces additional overhead, 
we further highlight that the sampling process has to be run only \textbf{once per dataset} even for two different models if they consume the same number of frames as input.
This feature further reduces the consumption of redundant computational power compared to those \textit{on-the-fly} sampling methods since they need to recalculated the duplicated sample process during every tuning stages, not to mention that the HDF5 file can be shared online with potential users and researchers to download.

\paragraph{Case Study}
We take the experiment using All-in-one on TGIF-QA as an example.
If using \textit{on-the-fly} uniform sampling, the training time per epoch is 52 min and the model takes 15 epoches to converge (780 min in total). 
As comparison, after applying our sampling methods, the training time per epoch reduces to 18 min per epoch (270 min in total) while the additional overhead to generate the .h5 file is 3 hour (180 min). 
The total time combining sampling and training and is $270+180=450$ min, much shorter than the implementation with on-the-fly sampling.

\section{Dataset Statistics}
We list the specifications of the datasets used in our evaluation process in~\Cref{tab:data_spec}.

\begin{table}[ht]
\resizebox{\linewidth}{!}{
    \begin{tabular}{c|c|c|c|c|c}
    \toprule
    Item & Split & MSVD & MSRVTT & TGIF & NExT \\
    \midrule
    \multirow{3}*{\#Video} & Train  & 1,200 & 6,513 & 37,089 & 3,870 \\
    ~ & Dev  & 250 & 497 & - & 570 \\
    ~ & Test  & 520 & 2,990 & 9,219 & 1,000 \\
    \midrule
    \multirow{3}*{\#Q\&A} & Train & 30,933 & 158,581 & 39,392 & 31,173 \\
    & Dev & 6,415 & 12,278 & - & 4,682 \\
    & Test & 13,157 & 72,821 & 13,691 & 16,189 \\
    \bottomrule
    \end{tabular}
}
    \caption{Statistics of the four QA datasets evaluated in this paper. The split row lists the number of corresponding items in train/dev/test set. Note TGIF-QA does not have a validation set.}
    \label{tab:data_spec}
\end{table}

\section{Hyperparameter Search}
In MDF, we run experiments on the sampled datasets with $\alpha \in \{2.3, 2.5, 2.7\}$.
In MIF, we first uniformly pre-sample 16 frames in all experiments, then we calculate question--caption matching score based on these sampled frames.
For all other hyperparameters (batch size, vocabulary size, learning rate, etc), we keep them same as original setting from their blogs or papers (for CLIP we adopt the same setting as GIT).

\end{document}